\documentclass[10pt,twocolumn,letterpaper]{article}

\usepackage[numbers]{natbib}
\usepackage{multicol}
\usepackage[named]{xcolor}
\usepackage{graphicx}
\usepackage{fancyhdr}
\usepackage[T1]{fontenc}
\usepackage{float}
\usepackage{stfloats}

\usepackage{subfigure}

\usepackage{wacv}
\usepackage{times}
\usepackage{epsfig}
\usepackage{graphicx}
\usepackage{amsmath}
\usepackage{amssymb}

\usepackage{tabu}
\usepackage{boldline}
\usepackage{multirow}
\usepackage{hhline}
\usepackage{array}

\definecolor{my_orange}{RGB}{255, 165, 0}

\usepackage[nolist]{acronym}

\long\def\invis#1{}  
\def\wacvPaperID{421} 

\wacvfinalcopy 
\ifwacvfinal
\def\assignedStartPage{9876} 
\fi

\ifwacvfinal
\usepackage[breaklinks=true,bookmarks=false]{hyperref}
\else
\usepackage[pagebackref=true,breaklinks=true,colorlinks,bookmarks=false]{hyperref}
\fi

\ifwacvfinal
\else
\fi

\begin{document}

\title{Seeing Through your Skin: Recognizing Objects with a Novel  Visuotactile Sensor}

\author{F. R. Hogan$^{1}$, M. Jenkin$^{1,3}$, S. Rezaei-Shoshtari$^{1,2}$, Y. Girdhar$^{1,4}$, D. Meger$^{1,2}$, and G. Dudek$^{1,2}$\\
$^1$Samsung AI Center Montreal, Canada, \hspace{2mm} $^2$McGill University, Canada \\
$^3$York University, Canada,  \hspace{2mm}
$^4$Woods Hole Oceanographic Institution, USA
}

\maketitle

\begin{abstract}

We  introduce  a  new  class  of  vision-based
sensor  and associated algorithmic processes that  combine  visual  imaging  with  high-resolution
tactile sending, all in a uniform hardware and computational architecture.  We demonstrate the sensor's efficacy for both
multi-modal object recognition and metrology.  Object  recognition is typically  formulated as an unimodal task, but by combining two sensor modalities we show that we can achieve several significant performance improvements.
%
This sensor, named the See-Through-your-Skin sensor (STS), is designed to provide rich multi-modal sensing of contact surfaces. Inspired by recent developments in optical tactile sensing technology, we address a key missing feature of these sensors: the ability to capture a  visual perspective of the region beyond the contact surface. Whereas optical tactile sensors are typically  opaque,  we present a sensor with a semitransparent skin that has the dual capabilities of acting  as  a tactile sensor and/or as a visual camera depending on its internal lighting conditions. This paper details the design of the  sensor, showcases its dual sensing capabilities, and presents a deep learning architecture that 
fuses vision and touch. We validate the ability of the sensor to classify household objects, recognize fine textures, and infer their physical properties  both through
numerical simulations and experiments with a smart countertop prototype.
\end{abstract}

\section{Introduction}

\begin{figure}[t]
\centering
\includegraphics[width=\columnwidth]{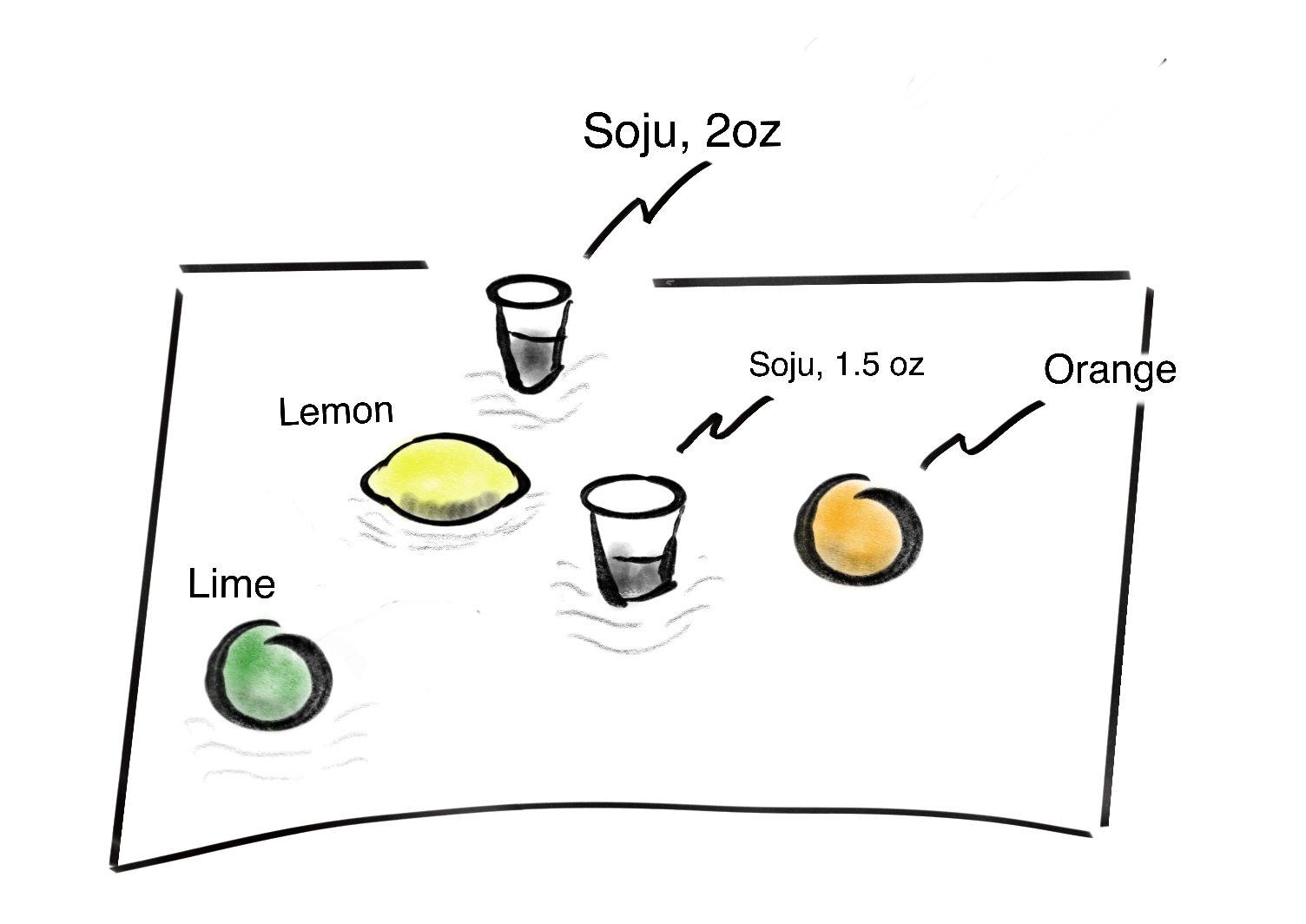}
\caption{The smart countertop scenario. How can a device recognize and characterize objects placed on a countertop, based on their visual appearance, their texture, and their mass?}
\label{fig:countertop}
\end{figure}

A core task in the development of smart homes and offices is enabling the environment with the ability to recognize and characterize objects placed within it. As an illustrative example, Fig. \ref{fig:countertop} shows a smart countertop that recognizes objects placed on its surface. This type of device finds applications in smart kitchens, interactive displays, and robotics.  
Central to this task is  the well-studied problem of object recognition. This paper introduces a novel sensing technology that utilizes an integrated tactile-vision sensor to perform object recognition and metrology tasks. The vast majority of efforts in object recognition are unimodal, relying on an individual sensing modality to perform the task. This can make certain recognition tasks  difficult as some  objects can appear very similar in one modality but are easily distinguished in another (e.g., lemons and limes have similar textures but different colors). This paper investigates how to enable and exploit multiple sensing modalities to optimize object recognition within the household domain. 

Humans and non-human primates are known to fuse tactile and visual information at a fundamental level~\cite{humanFMRI2011}. It is  intuitive for a human to mix visual tracking with touch sensing during  daily living, such as being able to look away from a bottle while pouring, after initially  localizing  the target. In automated perception, fusing vision and touch signals from distinct sensors has often shown increased performance compared with either individual modality~\cite{Quinn2014neuroscience}. Similarly, machine systems that combine vision and tactile information have shown enhanced performance over systems that utilize either vision or tactile cues alone~\cite{Corradi:2017}. One problem with integrating high resolution tactile and visual information is capturing the information within a common reference frame. 


Recent approaches to tactile sensing (e.g., GelSight~\cite{yuan2017gelsight}) include a camera that captures the deformations of a reflective soft surface as it contacts the world. This enables high resolution reasoning about contact geometry as well as slip and contact forces, but unfortunately light from the external world does not reach the camera due the sensor's skin opacity, preventing its use as a traditional vision sensor. Here, we augment the approach described in \cite{yuan2017gelsight} by making the deformable surface semitransparent. This enables the sensor  to capture both visual appearance and tactile properties simultaneously and from the same viewpoint. By equipping the sensor with  programmable internal lighting, this {\em See-Through-your-Skin (STS)\/} sensor provides a rich viewpoint of both the tactile and visual properties of the environment simultaneously. 

\begin{figure*}[t]
\centering
\includegraphics[width=1.\textwidth]{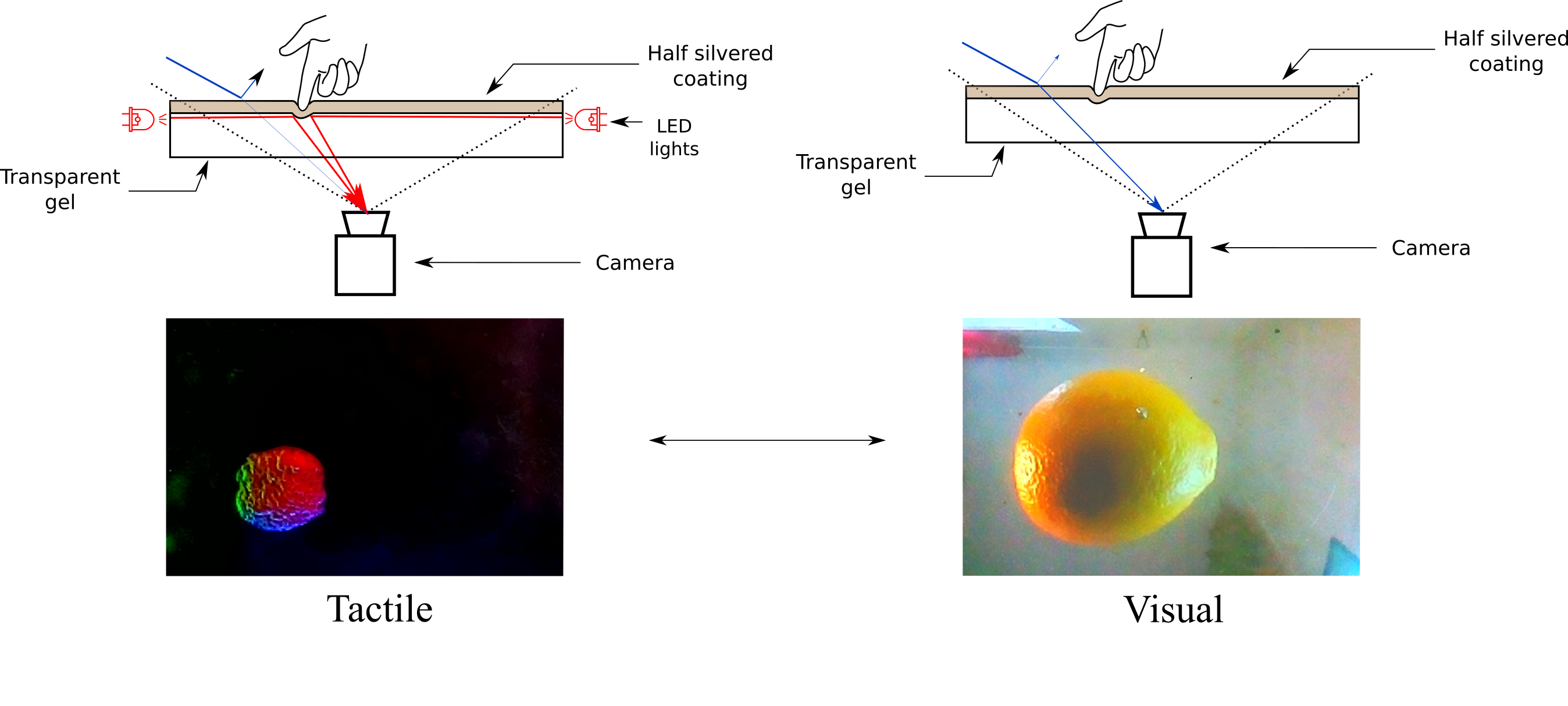}
\vspace{-30pt}
\caption{The two modes of the the STS sensor viewing the same object.  Through selective internal illumination, the surface of the gel can be made to be transparent (right), allowing the camera to view the outside world or opaque (left), where it works much like existing tactile sensors.  When the interior lighting levels are high relative to the outside world (left hand side) the surface becomes opaque where it delivers only tactile information. When the interior lighting level is low relative to the outside of the sensor (right hand side) the camera can view through the gel and recover the outside world. }
\label{fig:sts_both_modes}
\end{figure*}

\begin{figure*}[t!]%
\centering
 \subfigure[STS components]{%
 \includegraphics[width=.49\textwidth]{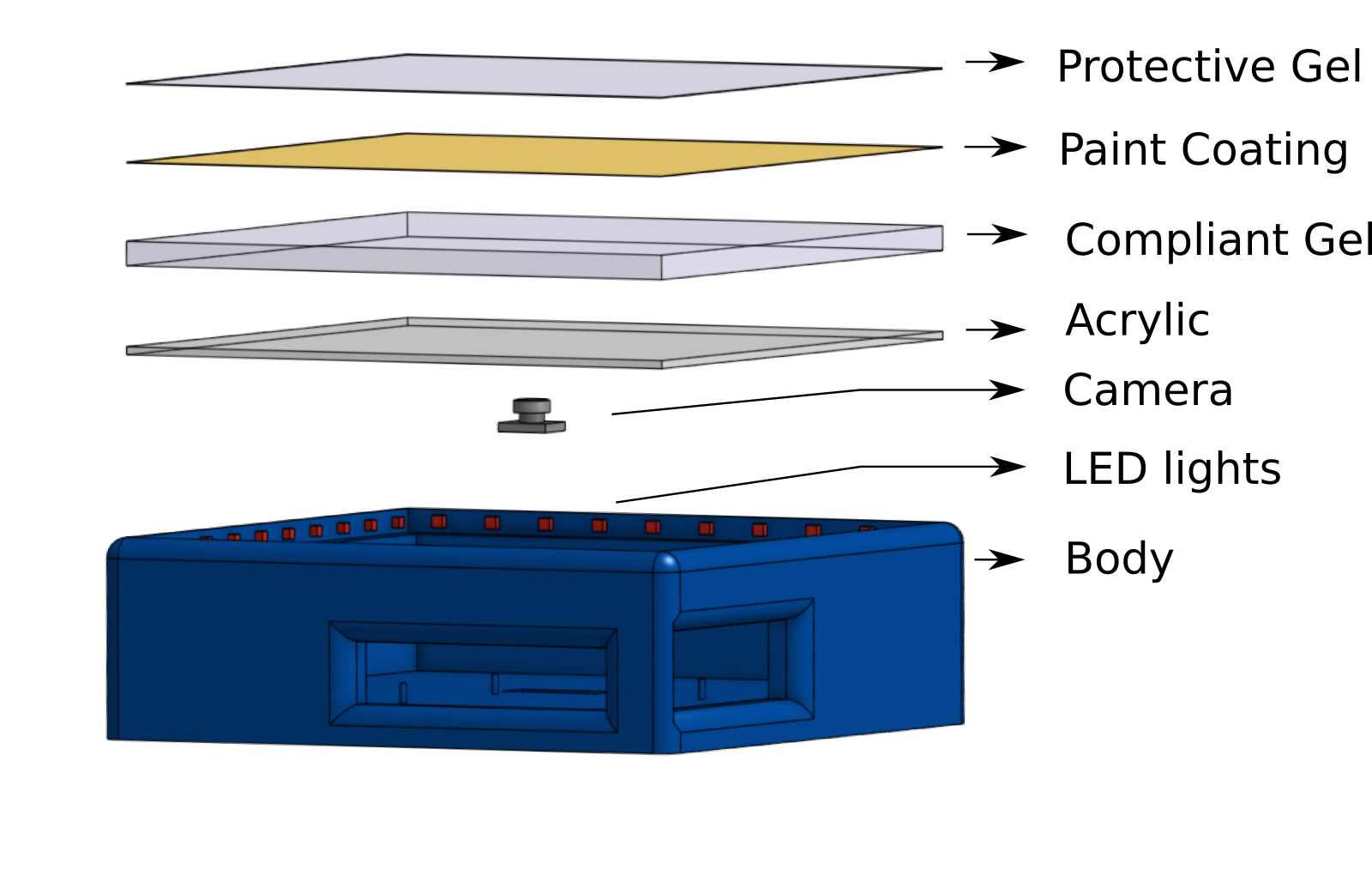}
 } 
 \label{fig:sts_components}%
 \hspace{5mm}
\subfigure[STS prototype]{%
 \includegraphics[width=.35\textwidth]{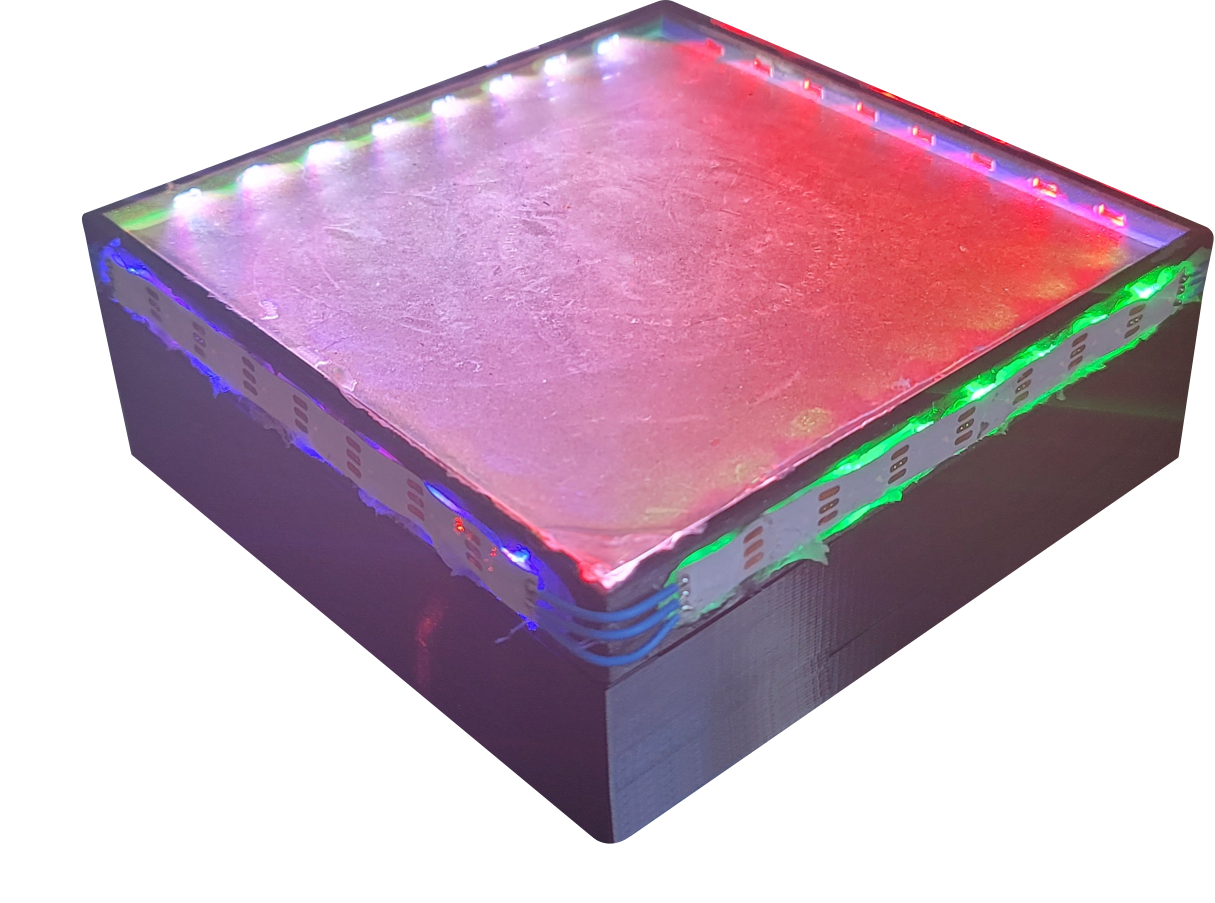}}
\caption{See-Through-your-Skin (STS) visuotactile sensor. The sensor is designed to provide simultaneous and high
resolution visual and tactile feedback by using a half-silvered paint coating over a compliant gel. By changing the internal
lighting condition of the STS sensor, the transparency
of the reflective paint coating of the sensor can be con-
trolled. 
 }
    \label{fig:sts_sensor}
\end{figure*}

The basic concept is illustrated in Fig.~\ref{fig:sts_both_modes}, which shows the multi-modal sensor  nature of the device and Fig.~\ref{fig:sts_components} which shows a prototype of the sensor and its components.  In this paper, we present a Deep Learning approach that exploits the multi-modal nature of the sensor to classify objects  by capturing the intrinsic correlations between both high-resolution sensory streams. Our experiments show that this approach can classify objects with high accuracy, and in particular, that objects whose properties confound one of the sensor modalities can still be recognized accurately in the joint visuotactile signal. 

The main contributions of this paper are:
\begin{itemize}
    \item \textbf{Design} of a novel visuotactile sensor, the See-Through-your-Skin (STS) sensor, that combines vision-based sensing with tactile feedback. By regulating the internal lighting of the sensor, we can control the type of feedback  collected by the sensor (tactile or visual).
\item \textbf{Perception} of multi-modal sensory streams using a Deep Learning framework. The network architecture fuses  vision  and  touch using dual stream convolutional signals to accomplish object recognition tasks, such as object classification and weight detection.
%

\item \textbf{Prototype} of a smart  countertop, a $15$cm $\times$ $15$cm surface sensorized with the STS sensor providing visuotactile feedback. 
\end{itemize}

We  validate the ability of the sensor  to recognize household objects and infer their physical properties both through in simulation and experimentally with a prototype version of the sensor. 

\section{Related Work}
This section outlines previous relevant research in the areas of tactile sensor design,  robotic applications, and multi-modal visual-tactile perception. 

\textbf{Tactile Sensing:}
Tactile sensors capture fundamental properties about the physical interaction between two objects, including the contact shape, texture, stiffness, temperature, vibration, shear and normal forces~\cite{Tiwana:12}. We refer the readers to \cite{Chi:18, KappassovTactileReview} for a comprehensive review of existing tactile sensing technologies. Optical tactile  sensors \cite{Shimonomura:19}   use a combination of a light source and a detector, typically a camera, to capture the contact interface. Current optical technologies that support compliant tactile sensing are constructed of an opaque interaction surface covering a transparent, typically gel-like or clear elastomer\cite{yuan2017gelsight}. Deformations of the opaque surface are monitored using a camera mounted underneath the surface while the transparent gel provides a physical structure to support the surface. In recent years, there have been a number of GelSight inspired tactile technologies. GelSlim   \cite{donlon2018gelslim} presents a slender and robust finger designed for robotic manipulation applications. DIGIT \cite{Lambeta2020DIGIT} present a miniaturized high-resolution tactile sensor designed for in-hand robotic manipulation. Omnitact \cite{padmanabha2020omnitact} is a curved high-resolution tactile sensor that integrates multi-directional cameras, and is designed to be used as a robot fingertip. FingerVision \cite{yamaguchi2016combining} integrates an internal camera facing a stretched transparent elastomer embedded with printed dots. While FingerVision  allows for multi-modal visual tracking feedback and tactile feedback, its tactile resolution is limited by the number of printed dots (approximately $30$) on the elastomer. In contrast, the STS design introduced in this paper offers a high resolution   a tactile resolution equivalent to that of the camera ($1640$ $\times$ $1232$) allowing for the recovery of richer feedback about the contact geometry and texture. 

\textbf{Robot Applications}
Tactile perception has long been studied as a method for robot arms and hands to acquire shape information, as related in a recent survey by Luo \emph{et al.} \cite{Luo2017}. The geometric information obtained by detecting contact points can be fused to complete a coherent shape estimate \cite{Allen1984Surface,Bierbaum2008,Ottenhaus2016,Yi2016Active,17-driess-IROS,Luo2016} of the manipulated object. Touch has also been used to perceive the environment at a global level, to recognize objects and understand their placement within the scene \cite{Pezzementi2011}. For touch-enabled robots, it is often of interest to compute the sequence of arm and hand motions that will best facilitate geometric understanding \cite{Pezzementi2011,MartinezHernandez2013,Yi2016Active,Sommer2014,17-driess-IROS}. Finally, as robots increasingly become able to manipulate the world in a dexterous fashion, tactile sensing has been used in an integrated fashion to guide tasks with persistent contact such as in-hand manipulation \cite{Bierbaum2008potential,Sommer2014,Lambeta2020DIGIT}.


\textbf{Multi-modal Perception:}
The senses of touch and vision are complementary, and have often been combined to form a multi-modal perception framework for understanding object geometry \cite{Allen1984Surface,Bjorkman2013,Ilonen2014Three,Wang20183D,GANDLER2020103433}. Most recently, deep representation learning has been shown as a powerful tool to extract shared geometric knowledge from touch and vision \cite{LimPRPA19}. These systems boost recognition performance when all information is available, and perhaps most interestingly, by training with both modes, the individual sensory streams perform better even if only a single information channel is available. In robotic manipulation, a coordinated eye-and-hand feedback system permits accurately tracking the manipulated object and regulating the applied contact forces  \cite{Lee2019,watkins2019multi}.

\section{Sensor Design}
\label{sec:sensor_design}

The STS is designed to provide simultaneous and high resolution visual and tactile feedback. The key features of the sensor are:
\begin{enumerate}
    \item \textbf{Multi-modal perception}. The STS sensor provides both visual and tactile feedback.
    \item \textbf{Controllable transparency}. By changing the internal lighting condition of the STS sensor, the transparency of the reflective paint coating of the sensor can be controlled. Additionally, the controllable lighting can be used to assist in the recovery of tactile information.
    \item \textbf{Collocated high-resolution sensing}. The sensor makes use of the same camera as a receptor for both the tactile and visual feedback. This result is two sensing signals that have the same point of view, frame of reference, and resolution.
\end{enumerate}

\begin{figure}[t!]%
\centering
 \includegraphics[width=.49\textwidth]{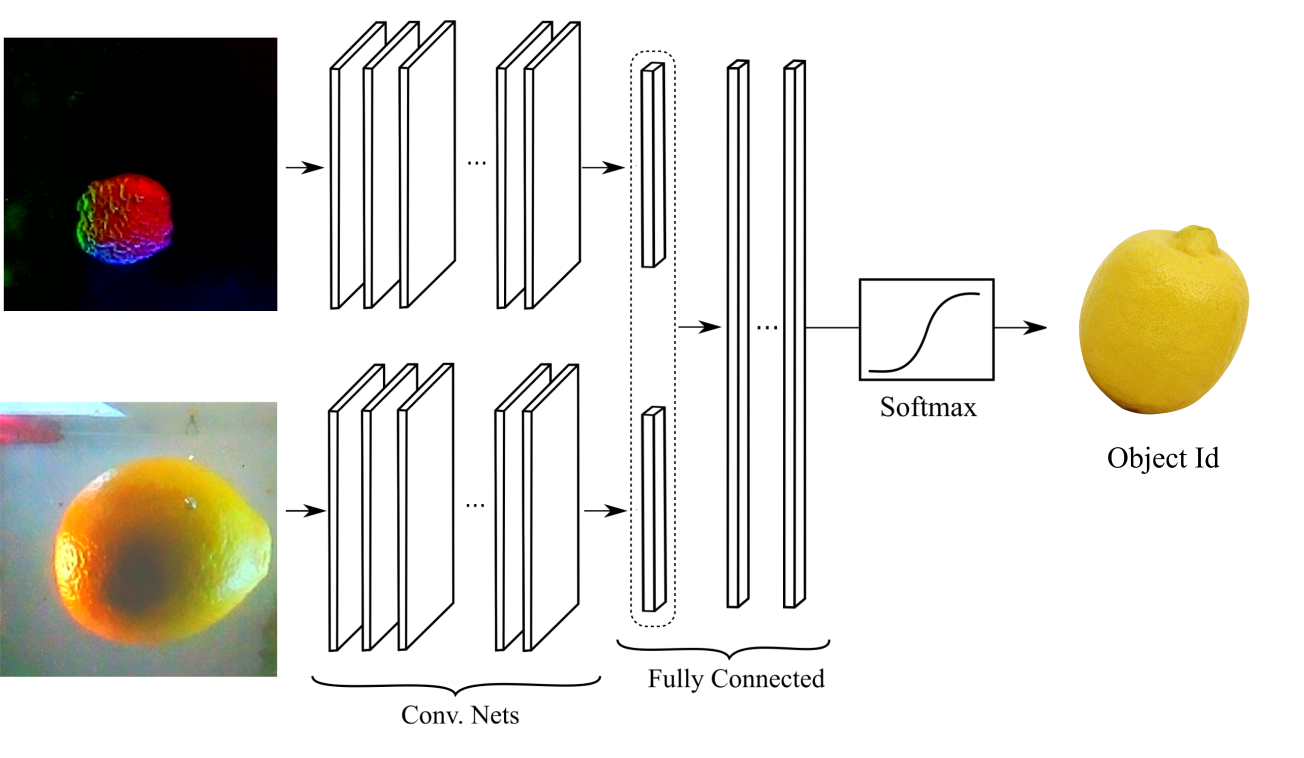}
 \label{fig:simulation_network}%
\vspace{-5pt}
\caption{Multi-modal network architectures. 
The tactile and visual images are processed  by independent convolutional neural networks, whose outputs are concatenated and fused into a fully connected network. 
}
\label{fig:network_architecture}
\end{figure}


\subsection{Background}
Optical tactile sensors   use a combination of a light source and a detector, typically a camera, to capture the contact interface. A property of many existing tactile sensors is that they are opaque and thus obscure the view of the object just at the critical moment prior to  contact. Current optical technologies that support compliant tactile sensing are constructed of an opaque interaction surface covering a transparent, typically gel-like or clear elastomer \cite{yuan2017gelsight}. Deformations of the opaque surface are monitored using a camera mounted underneath the surface while the transparent gel provides physical support. In order to improve the observability of the membrane deformation occurring during contact,  the  skin of compliant optical based tactile sensors is layered with an opaque layer of reflective material, which  has the side effect of obscuring the view to the external world.  Although the GelSight technology has been described as being able to ``see through your hands''~\cite{Coxworth:14}, this technology does not actually afford a visual modality as you cannot see through the sensor.


\subsection{See-Through-your-Skin (STS) Sensor}
The See-Through-your-Skin sensor (STS) uses   a half-silvered paint coating to allow for multi-modal sensing that is dependent on the lighting conditions. This setup, depicted in Figs. \ref{fig:sts_both_modes} and \ref{fig:sts_components}, behaves similarly as a ``one-way mirror'', also known as a ``transparent mirror''. When one side of the membrane is held brighter than the other, the membrane acts as an opaque mirror in the bright compartment and as a transparent window in the dark section. This phenomena is similar to the one  used in  police interrogation rooms, where a half-silvered glass window is used to separate the interrogation room from the observation room, both kept under different lighting conditions. By maintaining the observation room dark and the interrogation room bright, the same glass window appears as a mirror from the perspective of the suspect and as a clear window from the perspective of the observers.


The sensor, shown in Fig.~\ref{fig:sts_components},  is composed of two principal components:  a compliant skin and a rigid body. The compliant skin of the sensor comprises three components: a compliant gel, a reflective paint coating, and a protective gel. The compliant gel ensures that applied forces on the sensor result in physical deformations that bend the path of light as shown in Fig.~\ref{fig:sts_both_modes}. Similarly to \cite{donlon2018gelslim}, we use the  P-595 gel  by Silicones Inc. due to its optical clarity and its desirable stiffness properties. For the reflective paint coating, we use a reflective and translucent coating (Rust-Oleum 267727), commonly referred to as  ``mirror spray.'' This product is designed to be sprayed onto a clear glass sheet to convert it into an opaque mirror by successively applying $5$-$6$ coatings. By limiting the application to $2$-$3$ coatings, we can apply a thin reflective layer such that it  appears as translucent when the inside of the sensor is maintained dark relative to the exterior and opaque when maintained bright relative to the exterior. One challenge with employing reflective spray paint is its tendency to rapidly wear during physical interactions. For protection, we apply a thin layer of silicone coating on top of the reflective paint coating, by using a $50/50$ silicone/thinner mix, such that the silicone is liquid enough to be spread  evenly over the paint coating. During curing, the thinner evaporates, leaving a very thin layer to protect the paint coating without negatively impacting the sensor's  sensitivity.  

The sensor's  rigid housing includes a  sheet of acrylic to support the compliant gel.  
We capture physical interactions with the gel  using the  $160^{\circ}$ Variable Focus Camera Module for Raspberry Pi by Odeseven, due to its small form factor, its wild field of view, and its short focal length. 
To illuminate the sensor, we use the Neopixel LEDs by adafruit, due to their easily programmable interface. We use  a constant illumination pattern from the LEDs (clockwise per side: blue, red, white, green). Using different colors for the LEDs is motivated by previous work conducted in reconstructing the 3D contact geometry from RGB images using photometric stereo \cite{woodham1980photometric}. We control the LEDs and  stream the sensor's images at $90$Hz  onboard a Raspberry pi 3B+, located within the 3D printed body.

Figure~\ref{fig:sts_both_modes} shows the output of the STS while changing the internal lighting conditions of the sensor. As the illumination within the sensor increases, the half-mirrored membrane behaves more like an opaque mirror that renders a high resolution image of the contact geometry, effectively acting as a tactile sensor. As the light is decreased within the sensor, the light rays from outside the sensor can penetrate through the sensor and render a view of the external world, effectively acting as a visual sensor.




\begin{figure}[t]
    \centering
    \includegraphics[width=0.49\textwidth]{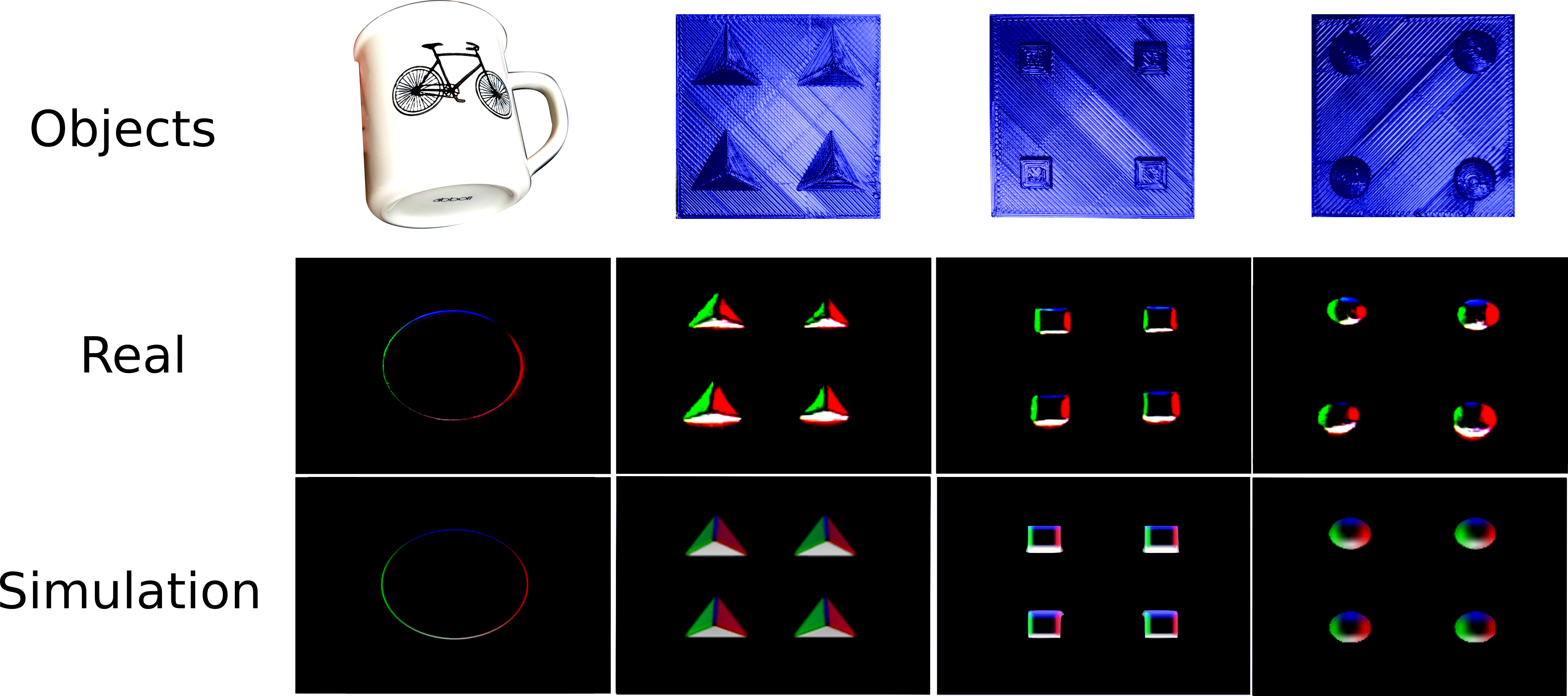}
    \caption{{Real vs. simulated comparison of STS tactile sensor}. The sensor computes surface deformations based on the contact forces obtained from the physics engine PyBullet. 
    }
    \label{fig:simulation_dataset}
\end{figure}

\subsection{Simulation}
\label{sec:simulation_setup}

We  present a visuotactile simulator that outputs a high resolution image of the contact geometry.  We leverage the simulator to test the sensor's ability to recognize objects, to explore the best way of encoding data from the sensor, and gain intuition regarding the importance of visual versus tactile information for this task.  The simulator maps the geometric information about the object in collision with the sensor via: 
\[
    \mathbf{I}(x, y) = \mathbf{R}(\frac{\partial f}{\partial x}, \frac{\partial f}{ \partial y}),
\]
where $\mathbf{I}(x, y)$ is the image intensity, $z = f(x, y)$ is the height map of the sensor surface, and $\mathbf{R}$ is the reflectance function modeling the environment lighting and surface reflectance \cite{yuan2017gelsight}.

%

The surface function $f$ is  obtained from the depth buffer of an OpenGL camera in PyBullet, which we  clip  to the thickness of the STS elastomer ($5$mm). To compute the surface normal at each point, we locate its adjacent points and calculate their principal axis using covariance analysis. Following \cite{gomesgelsight}, we  implement the reflectance function $\mathbf{R}$ using Phong's reflection model, which breaks down the lighting into three main components of ambient, diffuse, and specular for each RGB channel:
\begin{align}
    \mathbf{I}(x, y) &= k_a i_a + \nonumber \\
    & \sum_{m \in lights} k_d (\hat{L}_{m} \cdot \hat{N}) i_{m, d} + k_s (\hat{R}_m \cdot \hat{V})^\alpha i_{m, s}, \nonumber
\end{align}

where $\hat{L}_m$ is the direction vector from the surface point to the light source $m$, $\hat{N}$ is the surface normal,  $\hat{R}_m$ is the reflection vector computed by $\hat{R}_m = 2 (\hat{L}_{m} \cdot \hat{N}) \hat{N} - \hat{L}_{m}$, and $\hat{V}$ is the direction vector pointing towards the camera. Through extensive search and taking into account the suggested parameters in \cite{gomesgelsight}, we set the specular reflection constant $k_s$ to 0.5, the diffuse reflection constant $k_d$ to 1.0, the ambient reflection constant $k_a$ to 0.8, the shininess constant $\alpha$ to 5, and RGB channel of specular and diffuse intensities ($i_s$ and $i_d$) of each corresponding light source to 1.0. 

Tactile imprints are simulated with directional illumination orthogonal to the surface of the gel, using a constant illumination pattern from the LEDs (clockwise: blue, red, white, green). A comparison between the simulated and real world sensor tactile outputs is shown in Fig.~\ref{fig:simulation_dataset}.

The simulator uses a simple compliance model, which is approximated by modelling it with an array of springs (one per pixel), and solving for static equilibrium at each time step,  given the known contact geometry and reactive forces from the simulator.  

\section{Experimental Methods}
\label{sec:experimental_methods}
This section presents experimental methods used for data collection and evaluation of the results presented in Section~\ref{sec:results}.  We study the ability of the STS sensor to perform diverse object recognition tasks, including identifying i) objects drawn from the household domain ii) objects with subtle texture differences and iii) object with varying physical properties (mass). 

\subsection{Sensor Prototype}
\label{sec:sensor_prototype}
We prototype the STS sensor described in Section~\ref{sec:sensor_design} as a flat $15 \times 15$ cm surface, as  shown in Fig.~\ref{fig:sts_sensor}. This sensor is used to experimentally validate the sensor's ability to recognize and infer physical properties of objects. We prototype the sensor following the fabrication procedure described in Section~\ref{sec:sensor_design}, where we modulate the internal lights of the sensor at a frequency of $30$Hz. We separate the output of the sensor into visual/tactile data streams with an image resolution of $1640$ $\times$ $1232$. 


\subsection{Experiments}

We consider $3$ recognition scenarios evaluating the ability of the sensor to recognize diverse household objects  and infer their textural and physical properties.


\textbf{Household object recognition}. To evaluate  the ability of the STS sensor to recognize household
objects, we construct a dataset of visuotactile imprints taken with $10$ objects. We perform this analysis both in simulation and using the real world sensor.  In simulation, we draw $10$ objects  from the  3D ShapeNet dataset~\cite{chang2015shapenet}. The visuotactile dataset is comprised of 
$12$k simulated images ($600$ visual and $600$ tactile  per category), from which we use $70\%$ for training and hold out $30\%$ for validation.  For the real world experiments, we construct a dataset consisting of ten objects with subtle visual, textural, and mass differences (bottles). Each bottle is placed ``bottom down'' on the sensor $80$ times in different positions and orientations, where we capture a snapshot at the moment the bottle contacts the surface. We use $80$ percent of the data for training and the hold the remainder for validation.


\begin{figure*}[t!]
    \centering
    \includegraphics[width=\textwidth]{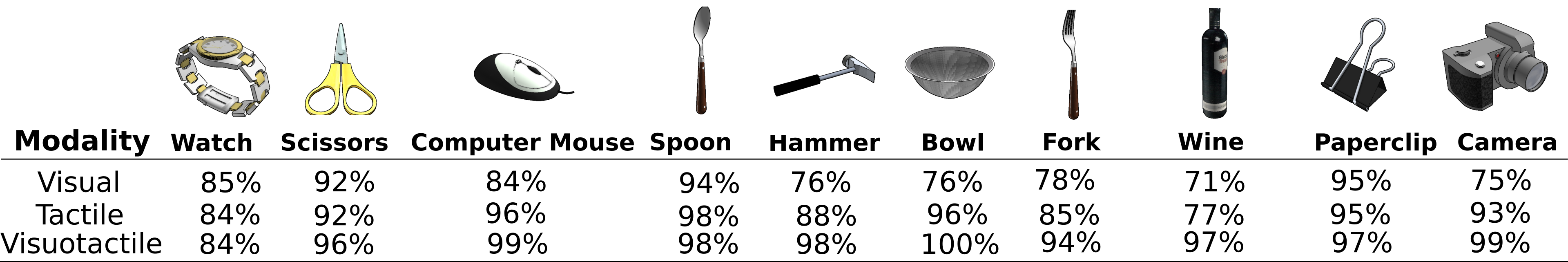}
    \vspace*{-10pt}
    \caption{Recall scores for each category of the simulated dataset. The scores are reported on the validation set after training to the 100\textsuperscript{th} epoch on the training set. Recall scores are reported for vision only, tactile only and visuotactile recognition.}
    \label{fig:simulation_recall_results}
\end{figure*}

\begin{figure}[b!]
    \centering
    \includegraphics[width=0.49\textwidth]{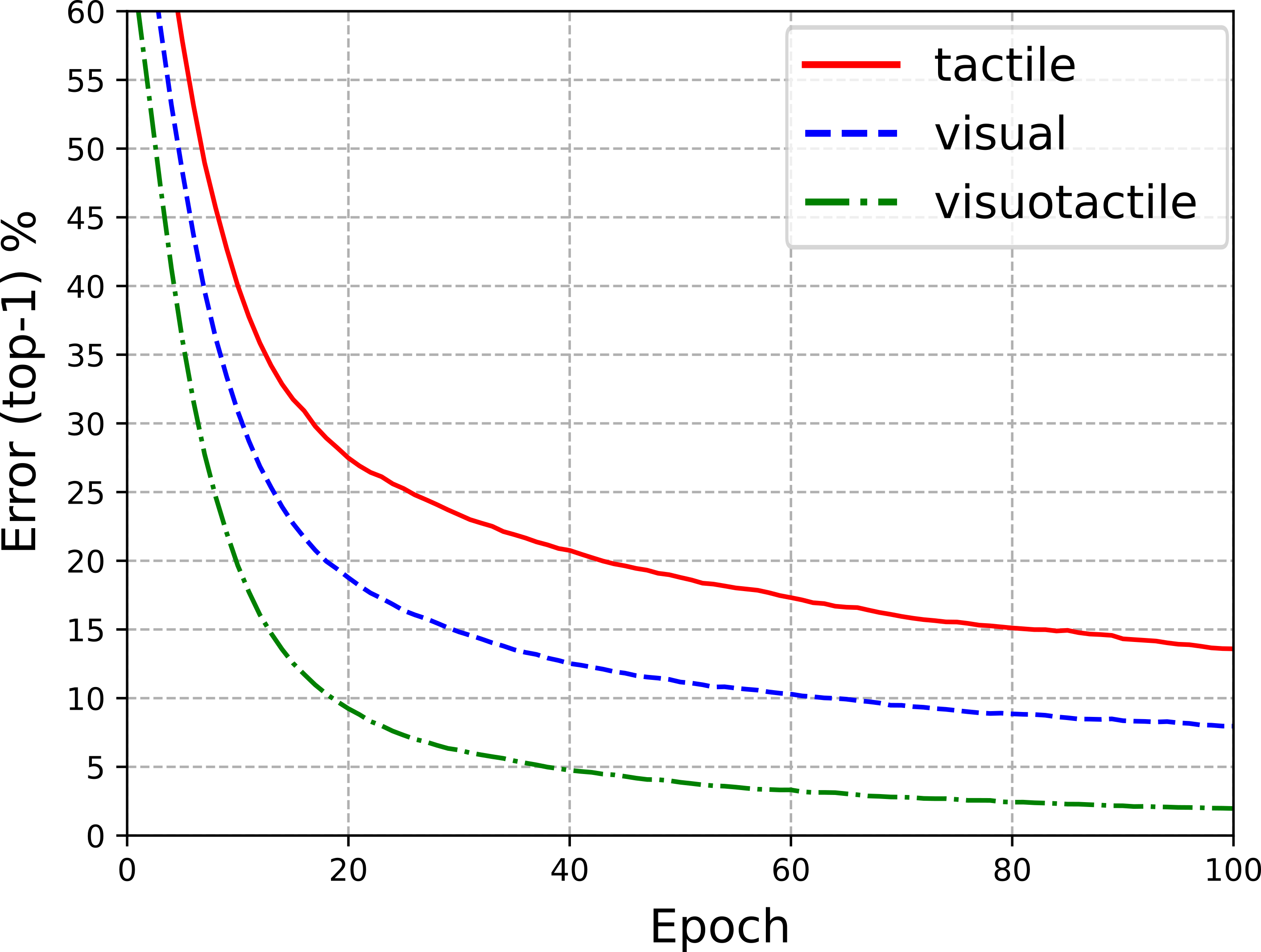}
    \caption{Learning curve for object recognition during training ResNet-50 with different modalities on the simulated dataset. The evaluation is measured on the validation set. Combination of both visual and tactile modalities results in faster training and higher accuracy.}
    \label{fig:simulation_results}
\end{figure}



\textbf{Texture recognition}. Beyond generic object recognition, we are interested in evaluating the ability of the STS sensor to recognize objects with small scale  textural differences. To this end, we construct a dataset consisting of $3$D printed objects, ranging from coarse textures to  fine textures, as shown in Fig.~\ref{fig:texture_recognition}. The objects are printed mate black, making it very difficult for visual-based classification. For each object, we collect a total of $100$ imprints for each object, where we use $80$ percent of the data for training and the hold the remainder for validation.  

\textbf{Metrology assessment}.  To evaluate the ability of the sensor to make quantitative disctinctions based on object physical properties, we consider the task of estimating the  amount of material in a container. 
We build a  dataset that is collected using a liquor bottle filled with $3$ fullness levels (empty: $446$g, half-full: $823$g, full: $1133$g). Each bottle is placed ``bottom down'' on the sensor $120$ times in different positions and orientations, where we capture a snapshot at the moment the bottle contacts the surface. We use $80$ percent of the data for training and the hold the remainder for validation.

\section{Results}
\label{sec:results}

This section considers the task of recognizing  objects based on their visual and physical properties, as detailed in Section~\ref{sec:experimental_methods}.
We demonstrate the capability of the STS sensor to identify: 1) diverse objects drawn from the household domain, 2)  objects with subtle textural differences and 3) distinctions based on metrology (i.e.\ differentiating bottles based on the different amounts of liquids in them). 

\begin{figure*}[t!]
    \centering
    \includegraphics[width=\textwidth]{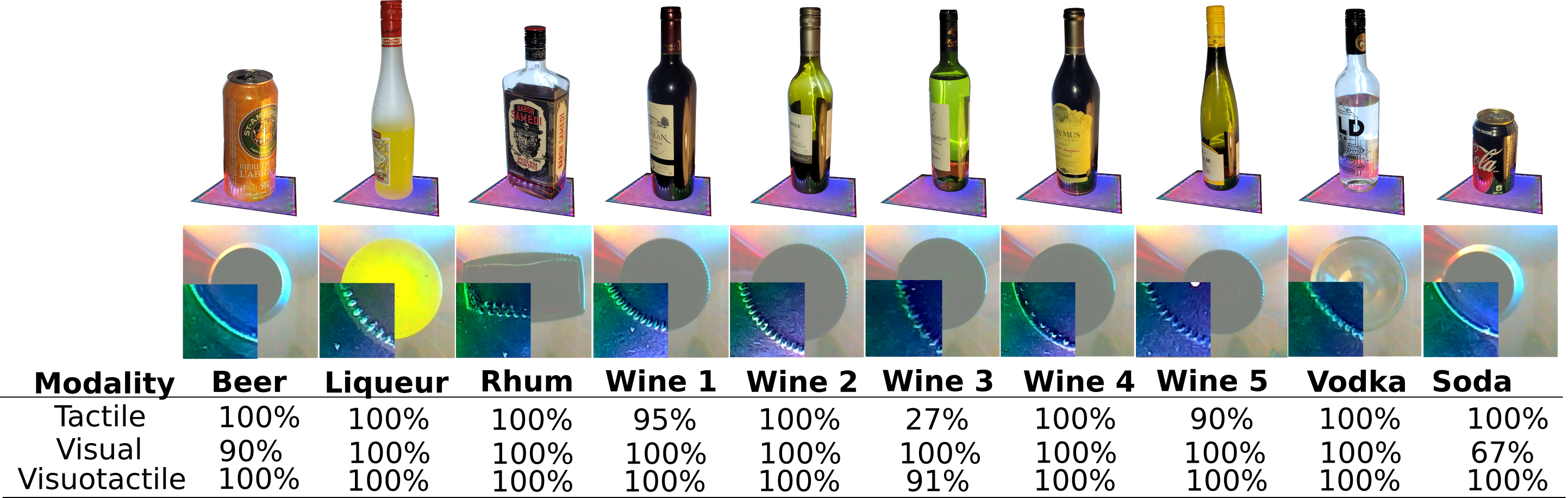}
    \vspace*{-10pt}
    \caption{Bottle classification with the STS. The bottom row shows sensor views of the corresponding bottle shown in the top row with recognition accuracy below each view. The scores are reported on the validation set after training to the 100\textsuperscript{th} epoch on the training set. Recall scores are reported for vision only, tactile only and visuotactile recognition.  The bottom left of each image shows a blownup of the tactile imprint of each bottle. Note that the sensor does not ``see'' the actual bottles, these are only included here for ease of exposition.}
    \label{fig:bottle_classification}
\end{figure*}

\subsection{Household Object Recognition}
\label{sec:simulation_results}

Our first experiment involves training our multi-modal learning framework, shown in Fig.~\ref{fig:network_architecture}, where the task is to distinguish between objects based on their category labels. This is a well-known perceptual task, and often the first step in automation applications. We are particularly interested in studying the significance of utilizing both the visual and tactile modalities from our unique sensor. \textbf{Can a classifier benefit from both inputs? Is the data quality from the sensor appropriate to obtain high classification accuracy? Is our deep architecture suitable for these inputs?} 

To answer these questions, we trained and validated learned models in a variety of settings.  Each scenario in our training and validation sets includes all of the possible modalities, where we  sub-select  the visual signal, only the tactile reading, or both modalities combined  to create three separate learning tasks. Using these three tasks, we compare the classification performance of single-modal and multi-modal classifier networks, using the network architectures depicted in Fig.~\ref{fig:network_architecture}. We use ResNet-50 \cite{he2016deep}, trained with the Adam \cite{kingma2014adam} optimizer for $100$ epochs with a learning rate of $1\mathrm{e}\scriptstyle{-}\displaystyle{4}$ and a batch size of $100$. Due to the limited number of real-world data, we pretrain the ResNet-50 networks on the ImageNet dataset. 

The simulation results in Fig.~\ref{fig:simulation_results} show the progress of learned models evaluated on the validation set for the different modalities. While both modalities achieve relatively good performance by themselves (visual: $11.25\%$ error, tactile: $16.92\%$ error), the overall accuracy is significantly improved (visuotactile: $3.12\%$ error) by fusing both modalities together using a dual stream convolutional network architecture as shown in Fig.\ref{fig:network_architecture}. Figure \ref{fig:simulation_recall_results} reports the classification recall scores for each object individually at the end of the 100\textsuperscript{th} epoch. Interestingly, as visual and tactile modalities provides independent information, the accuracy obtained using both modalities can significantly improve upon that obtained by any single modality. For example, the wine bottle's recall score improves from $71\%$ tactile and $77\%$ visual, to $97\%$ visuotactile. 

While the simulation-based analysis provides support for the value of our multi-modal sensor, we verify the quality of information provided by the real device, under realistic lighting and physical conditions. To this end, we evaluate the ability of the real-world STS prototype to differentiate household objects, where the task is to distinguishing different instances of object within the same category, in this case a variety of bottles. Sample sensor input signals are shown in Fig.~\ref{fig:bottle_classification} (bottom row) for the input bottles shown in Fig.~\ref{fig:bottle_classification} (top row). Note that the sensor does not see these bottles and classification is based only on the imaging of the bottom of the bottles as shown in Fig.~\ref{fig:bottle_classification}.  This problem is of particular relevance for study with a visuotactile sensor as bottles often have a common profile and overall shape, differing primarily in fine-grained markings on the   bottom surface that are difficult to detect visually. We trained and validated the STS on a dataset of real-world sensory readings that capture ten different representative  bottles with similar shapes and subtle textural differences, as described in Section~\ref{sec:experimental_methods}. Figure~\ref{fig:bottle_classification} shows accuracy scores obtained by our trained models. Results show that both visual and tactile inputs contain distinctive information useful to distinguish differen object categories. However, both tactile and visual individual modalities struggle with specific object instances.  With only tactile information, the classifier misclassifies wine bottle $3$, that it often confuses with wine bottle $1$ and $5$  given their similar tactile imprints (see supporting documentation for confusion matrices). With only visual information, the network properly distinguishes between the wine bottles based on  subtle color and shape differences. Conversely, the visual network has difficulties recognizing the \textit{coke} can, that is often mistaken for the \textit{beer} can, both having similar visual perspectives from below. Interestingly, the tactile network is able to differentiate between both instances, presumably due to their different mass. The ability of the STS sensor to make quantitative assessments is further explored in Section~\ref{sec:measurement_results}. When combining both modalities, the multi-modal network achieves  high recognition rates,  with all bottles achieving above $91\%$  accuracy scores,  overcoming the limitations of each  modality individual. \textbf{The multi-modal information provided by our sensor is significantly helpful for the task of recognizing diverse objects}.


\subsection{Texture Recognition}
\label{sec:texture_recognition}

\begin{figure*}[t!]
    \centering
    \includegraphics[width=.99\textwidth]{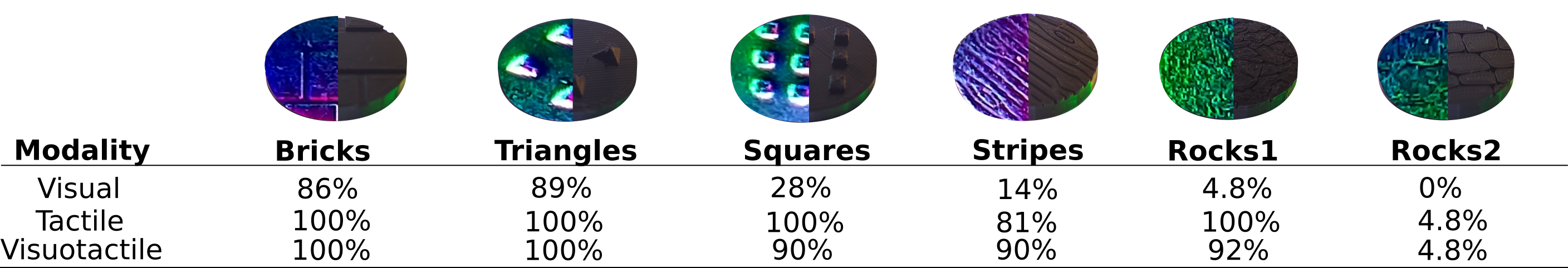}
    \caption{Texture recognition of $6$ objects with varying degrees of texture coarseness.  The scores are reported on the validation set after training to the 100\textsuperscript{th} epoch on the training set. Recall scores are reported for vision only, tactile only and visuotactile recognition.  Results show that  tactile feedback is required to accurately differentiate most object types, as the textural differences are too fine to be visually noticeable. }
    \label{fig:texture_recognition}
\end{figure*}

In this section, we investigate the ability of the STS sensor to recognize textures, by classifying $6$ objects with varying degrees of texture coarseness shown in Fig.~\ref{fig:texture_recognition}. For this task, we   ask: \textbf{can the STS sensor differentiate between fine texture differences?}

In Fig.~\ref{fig:texture_recognition}, we show the classification results on a test set of $6$ difference object textures of various levels of difficulty. We show the results for each mode individually, as well as combined. Results show that, as expected, visual feedback cannot accurately differentiate most object types, as the textures are too fine to be visually noticeable. The tactile modality, visualized and overlayed in Fig.~\ref{fig:texture_recognition},  augments the ability of the perceptual system to visualize the object textures, and is able to accurately distinguish between $5$ out of the $6$ objects with over $80\%$ accuracy. In the case of the object on the bottom right, the tactile sensor is unable to properly identify it due to the  downsampling of the image to the resolution of $(224, 224)$ used by the Resnet-50 architecture. Importantly, the combined visuotactile network achieves equally good performance as the tactile sensor, showcasing \textbf{the ability of the tactile sensor to differentiate between objects with fine textural differences.}

\begin{figure}[b!]
    \centering
    \includegraphics[width=.49\textwidth]{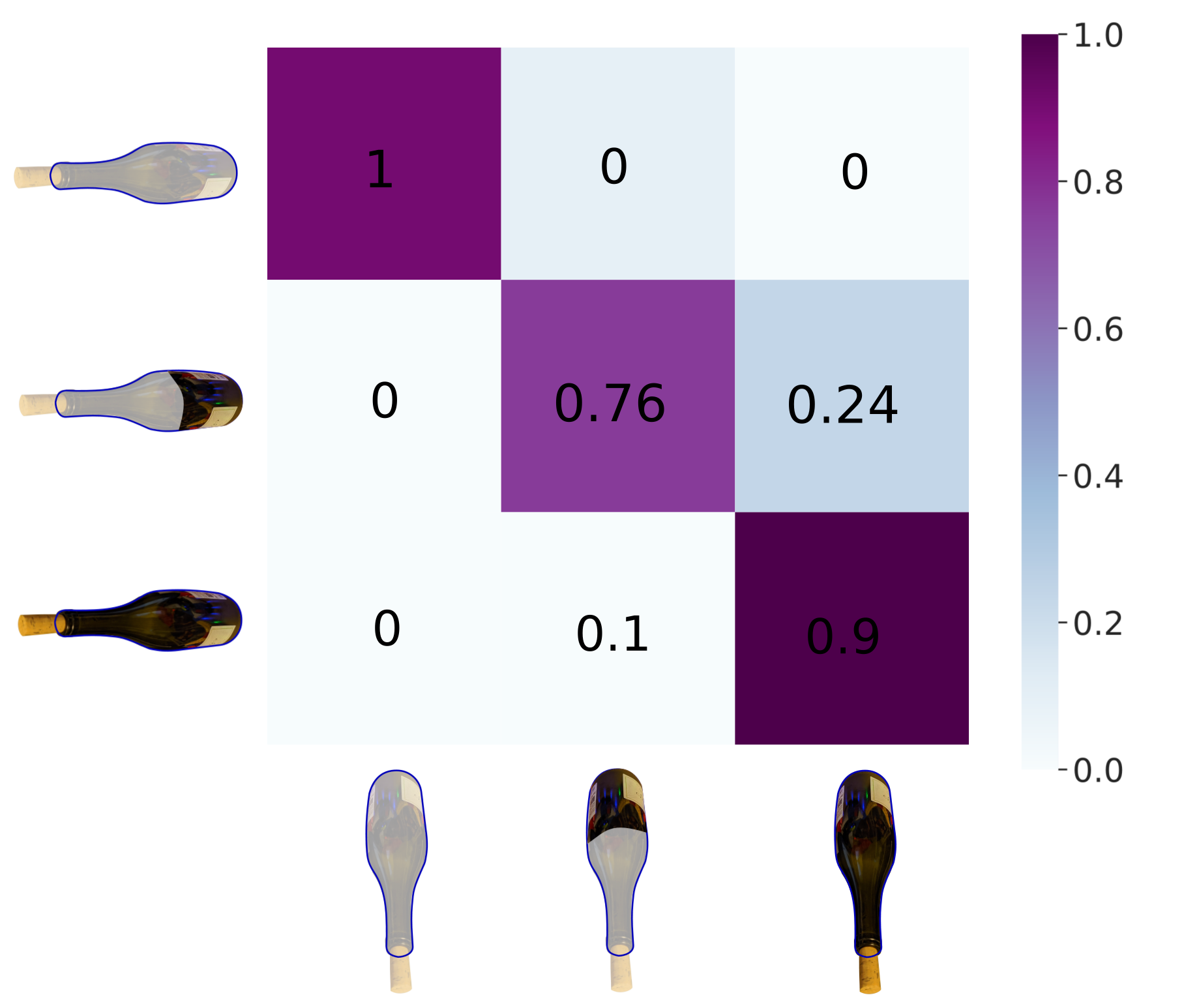}
    \caption{Confusion matrix for inference of bottle contents. Rows and columns represent true and predicted classes, respectively. }
    \label{fig:metrology_confusion}
\end{figure}

\subsection{Metrology Assessment}
\label{sec:measurement_results}

Our final experimental validation involves using the STS to make simple quantitative distinction between objects of different masses. This task would be useful on a smart counter top, for example, to infer the amount of liquid contained in bottles placed on the surface, which is completely unobservable in visual imagery in some opaque bottles. The problem of measuring liquid levels is a common problem in both the household kitchen as well as commercial restaurants and bars. \textbf{Can our multi-modal sensor enable inference of the fullness level of bottles placed on its surface?}

The objective here is not to compete merely with a conventional scale, but to exemplify and evaluate the potential for quantitative assessment combined with the other modalities of the sensor. Specifically, we varied the ``fullness'' of bottles by considering $3$ fullness levels (empty: $446$g, half-full: $823$g, full: $1133$g).  The training procedure follows the details described in Section \ref{sec:simulation_results} and uses the network architecture  depicted in Fig. \ref{fig:network_architecture}. 
Figure~\ref{fig:metrology_confusion} summarizes the confusion matrix for this model. As expected the cross-modal network achieves higher accuracy that a model trained with pure visual information, with the accuracy results for each modality is related in Table 
Interestingly, when the model incorrectly labelled a bottle, the most likely label was a bottle with the next closest volume. While a traditional scale would be  more accurate, these preliminary results to verify \textbf{our multi-modal sensor enables a reasonable ability to quantitative assesments of object physical properties}.



\section{Discussion and Future Work}
In this paper, we describe the principles, design, algorithmic foundation and evaluation of a new kind of sensor that simultaneously provides tactile information and visual appearance.  We sketch how it can be used for both categorical and quantitative assessment of object properties.

Integrating tactile and visual data into a common framework provides an effective signal for object recognition as well as quantifying physical properties of  objects. The sensor developed and prototyped in this paper  provides a combined visuotactile signal that eliminates the need to install parallel sensors with the advantage of a common reference frame. The STS is inexpensive, compliant, and leverages recent advances in vision-based tactile sensors. It provides a visual signal by replacing the opaque surface of sensors such as GelSight with a semi-transparent surface whose transparency nature can be modulated as necessary, but without which such modulation can obtain a combined tactile-visual signal. Additionally, as shown in this paper, the lighting conditions of the sensor can be modulated such as to obtain a single image that fuses both visual and tactile information and is sufficient to distinguish objects and to characterize their weight.

\begin{figure}[t!]
    \centering
    \includegraphics[width=.4\textwidth]{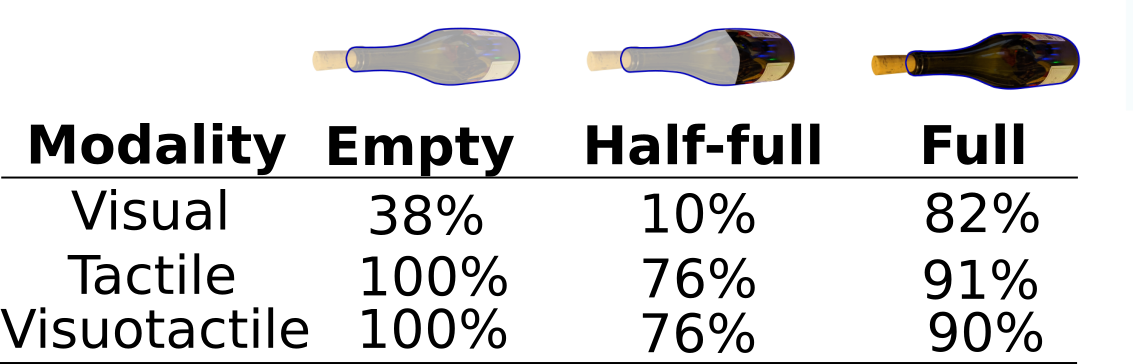}
    \caption{Bottle weight classification with the STS sensor. The scores are reported on the validation set after training to the 100\textsuperscript{th} epoch on the training set. Recall scores are reported for vision only, tactile only and visuotactile recognition. }
    \label{fig:metrology_ablation}
\end{figure}


In the longer term, we anticipate deploying the STS technology in larger scale surfaces capable of displaying information, as well as acting as an input medium and developing non-planar surfaces for specific applications (such as a smart eye-finger hybrid).



\bibliographystyle{unsrt}
\bibliography{egbib}

\end{document}